\documentclass[letterpaper]{article} % DO NOT CHANGE THIS
\usepackage{aaai24}  % DO NOT CHANGE THIS
\usepackage{times}  % DO NOT CHANGE THIS
\usepackage{helvet}  % DO NOT CHANGE THIS
\usepackage{courier}  % DO NOT CHANGE THIS
\usepackage[hyphens]{url}  % DO NOT CHANGE THIS
\usepackage{graphicx} % DO NOT CHANGE THIS
\usepackage{natbib}  % DO NOT CHANGE THIS AND DO NOT ADD ANY OPTIONS TO IT
\usepackage{caption} % DO NOT CHANGE THIS AND DO NOT ADD ANY OPTIONS TO IT
\usepackage{algorithm}
\usepackage{algorithmic}
\usepackage{newfloat}
\usepackage{listings}
\DeclareCaptionStyle{ruled}{labelfont=normalfont,labelsep=colon,strut=off} % DO NOT CHANGE THIS
\floatstyle{ruled}
\newfloat{listing}{tb}{lst}{}
\floatname{listing}{Listing}
\usepackage{booktabs}

\usepackage{xspace}
\newcommand{\Small}{\textit{Small}\xspace}
\newcommand{\Base}{\textit{Base}\xspace}

\newcommand{\Mse}{Dist-MPNet-CzEng\xspace}
\newcommand{\CommercialMse}{Dist-MPNet-ParaCrawl\xspace}
\newcommand{\RetroMAE}{RetroMAE-Small\xspace}

\title{Some Like It Small: Czech Semantic Embedding Models for Industry Applications}

\author{
    Jiří Bednář, Jakub Náplava, Petra Barančíková, Ondřej Lisický
}
\affiliations{
    Seznam.cz, Prague, Czech Republic\\
    \url{{jiri.bednar, jakub.naplava, petra.barancikova, ondrej.lisicky}@firma.seznam.cz}
}

\usepackage{eso-pic}

\begin{document}

\AddToShipoutPictureBG*{%
  \AtPageUpperLeft{%
    \hspace{0.5\paperwidth}%
    \raisebox{-\baselineskip}{%
      \makebox[0pt][c]{\textbf{This paper was accepted to IAAI 2024. Please reference it instead once published.}}
}}}

\maketitle

\begin{abstract}
This article focuses on the development and evaluation of~\Small-sized Czech sentence embedding models. Small models are important components for real-time industry applications in resource-constrained environments. Given the~limited availability of labeled Czech data, alternative approaches, including pre-training, knowledge distillation, and unsupervised contrastive fine-tuning, are investigated.
Comprehensive intrinsic and extrinsic analyses are conducted, showcasing the competitive performance of our models compared to significantly larger counterparts, with approximately 8 times smaller size and 5 times faster speed than conventional \Base-sized models. To promote cooperation and reproducibility, both the models and the evaluation pipeline are made publicly accessible.
Ultimately, this article presents practical applications of the developed sentence embedding models in~Seznam.cz, the Czech search engine. These models have effectively replaced previous counterparts, enhancing the overall search experience for instance, in organic search, featured snippets, and image search. This transition has yielded improved performance.

\end{abstract}

\section{Introduction}

In recent years, the field of natural language processing (NLP) has experienced remarkable progress. One key contributing factor to this progress is the development of more sophisticated distributed text representation, i.e., word and sentence embeddings \cite{devlin2018bert,liu2019roberta,reimers2019sentence}. 

In this work, our primary focus lies in the development and evaluation of sentence embeddings. Sentence embeddings are instrumental in various NLP tasks such as facilitating efficient information retrieval, sentiment analysis, machine translation, question answering, or providing interpretability. The importance of quality sentence embeddings simply cannot be overstated.

While sentence embeddings are becoming more sophisticated and capable of capturing richer semantic and syntactic information, there is a trade-off between embedding size and computational efficiency. In industry applications, the choice of an embedding model should be carefully evaluated. Larger embeddings may offer better representation of~sentence semantics, but they can significantly increase computational complexity and inference times, which may not be practical in real-time applications or resource-constrained environments.

Furthermore, it is essential to acknowledge that much of~the research and development in sentence embeddings has predominantly focused on English language, both in supervised and unsupervised settings. This creates challenges and limitations when applying these models to other languages, where data availability, linguistic characteristics, and semantic structures may differ significantly from English.

This paper aims to address the existing gap by focusing on~the~development of \Small\footnote{We follow the naming convention of~\citet{clark2020electra} in which \Small models refer to BERT-based models with roughly 14 million parameters, 12 encoder layers, and have a hidden size of 256. \Base models comprise circa 110 million parameters, 12 encoder layers and have a hidden size of 768.} sentence embeddings specially tailored for real-time industry applications in the Czech language. Nevertheless, the underlying techniques and evaluation can be adapted and applied to other languages as well. The research and strategies presented in~the paper serve as~a~valuable foundation for enhancing sentence embeddings, with the goal of optimizing computational efficiency without sacrificing their representational capacity and performance.

The main contributions of our paper are as follows:
\begin{enumerate}
    \item \textbf{\Small Czech models training}: We trained and evaluated multiple \Small Czech BERT \cite{devlin2018bert} based models for sentence embeddings. Despite being approximately 8 times smaller and 5 times faster compared to conventional \Base models, the \Small models exhibit competitive performance in diverse downstream tasks.
    \item \textbf{Czech sentence embeddings evaluation}: We conducted a thorough evaluation of existing Czech sentence embeddings, both intrinsically and extrinsically. This evaluation provides valuable insights into the effectiveness and applicability of different embeddings in various NLP tasks.
    \item \textbf{Evaluation pipeline}: To ensure transparency and facilitate the reproducibility and verification of our results, we have made our evaluation pipeline openly accessible. 
    \item \textbf{Public release of models}: Our developed models are publicly available under the CC-BY-4.0 and CC-BY-NC-4.0 licenses, depending on the training data used. This allows other researchers and practitioners to utilize and build upon our work in their own projects, fostering collaboration and advancement in the NLP community.
\end{enumerate}

All the developed models and the evaluation pipeline are available at the following link:
\begin{center}
\textbf{https://github.com/seznam/czech-semantic-embedding-models}\\
\end{center}

\section{Related Work}
Recently, two prominent methodologies, cross-encoder (CE) and bi-encoder (BE) \cite{reimers2019sentence}, have emerged as leading approaches in 
comparison of text pairs. Both methods encode sentence pairs and have demonstrated significant advancements in capturing the semantic meaning and relationships between sentences.

The CE approach entails encoding a pair of sentences together into a joint representation. By considering the interaction between the sentences, CE excels at capturing fine-grained semantic nuances and understanding complex relationships between text pairs \cite{qu2020rocketqa}. On the other hand, the BE approach takes an independent encoding approach \cite{zhao2022dense} -- it generates a separate embedding for each sentence, allowing for greater flexibility and computational efficiency. BEs are well-suited for large-scale applications like semantic search \cite{reimers2019sentence}, where the goal is to find the most similar sentence or document to a given query.  

While CEs excel at providing rich semantic information, BEs can partially achieve similar performance by integrating multiple text representations, as demonstrated in ColBERT \cite{khattab2020colbert} or MADRM \cite{kong2022multi}. %, which showed improvements in the e-commerce domain. 
However, the use of multiple representations can significantly increase the index size, making it impractical for certain applications. %Strategies introduced in 
ColBERTer \cite{hofstatter2022introducing} aims to mitigate this issue, thereby promoting wider adoption of~single-representation BE.

Improving the performance of sentence embedding models often relies not only on refining task-specific architectures but also on utilizing rich labeled English datasets. 
The Sentence Transformers\footnote{https://www.sbert.net} framework is widely used for comparing sentence embeddings and provides a range of models, including \textit{all-mpnet-base-v2}, fine-tuned with a contrastive objective \cite{song2020mpnet}, and \textit{all-MiniLM-L6-v2}, which provides high-quality embeddings while being computationally efficient. These models are trained on diverse datasets, covering over 1 billion sentence pairs.

Regrettably, there are no similarly extensive labeled datasets available for the Czech language. As a result, alternative strategies are necessary to optimize model performance. %Furthermore, the focus of training is often on \Base or even larger variants, making deployment challenging within the industry. 
One effective approach is knowledge distillation, where the excellent performance of larger, more complex models can be utilized to train smaller student models \cite{hinton2015distilling}. In this context, a CE can be employed as a teacher to train a BE \cite{qu2020rocketqa}. Interestingly, BE and CE can also be learned simultaneously in an unsupervised manner using the Trans-Encoder \cite{liu2021trans}, where one component generates pseudo-labels to update the other, eliminating reliance on labeled data. To address the scarcity of language-specific datasets, one possible solution is multilingual distillation \cite{reimers2020making}, where a strong teacher model, trained in English for example, is used to train a student model in the target language. The main advantage of this approach is that the student model learns representations for both languages. However, this training process requires a bilingual dataset whose quality directly impacts student model performance.

To further enhance model performance, an alternative strategy involves refining pre-training methods. One such approach involves leveraging autoencoders to compress text data into the “CLS” token using an encoder and then using a decoder to reconstruct the original text. This technique has been demonstrated in models such as Condenser \cite{gao2021condenser}, TSDAE \cite{wang2021tsdae}, and coCondenser \cite{gao2021unsupervised}. However, %it is worth noting that employing 
a strong decoder may negatively impact sequence representation quality \cite{lu2021less}. Recent approaches such as SimLM \cite{wang2022simlm} and RetroMAE \cite{liu2022retromae} address this issue by adopting shallow decoders with limited past context access and enhanced decoding mechanisms.

Another approach to enhancing text representation is through contrastive learning. A successful implementation of this approach is SimCSE \cite{gao2021simcse}. In SimCSE, authors applied a simple dropout mask as noise to the input text, creating positive pairs along with in-batch negatives. Despite its simplicity, this implementation has yielded surprisingly good results and has inspired numerous similar approaches, further improving model performance. Notable examples include DiffCSE \cite{chuang2022diffcse}, InfoCSE \cite{wu2022infocse}, and the recently introduced RankCSE \cite{liu2023rankcse}, which ensures ranking consistency between text pairs.

The effective utilization of pre-training strategies, such as employing autoencoders or contrastive learning, as well as incorporating knowledge distillation, can yield substantial improvements in capturing semantic meaning within text and consequently enhance the performance of models for languages that lack labeled datasets.

\section{Models}

Three distinct methodologies were employed in our work to train effective \Small bi-encoder
models and generate high-quality sentence embeddings: Auto-encoder training, unsupervised contrastive fine-tuning, and multilingual distillation. The primary advantage of these approaches is their independence from large supervised datasets, which are often unavailable for low-resource languages like Czech.

\subsection{RetroMAE}
The recent retrieval-oriented pre-training method, RetroMAE \cite{liu2022retromae}, was used in combination with \Small configurations of BERT as an encoder. The setup included 12 layers, a hidden size of 256, and a WordPiece tokenizer with vocabulary size of 57,226, incorporating both Czech and English tokens. The original authors' code and configuration were utilized, but the training hyperparameters were adjusted to accommodate the smaller model size. Asymmetrical masking ratios were applied: 30\% for the encoder and 50\% for the decoder. The Adam optimizer \cite{kingma2014adam} with decoupled weight decay regularization \cite{loshchilov2018decoupled} (AdamW) was used in the training process with a learning rate of 5e-4, along with a cosine scheduler \cite{loshchilov2017sgdr} with a linear warmup of 10\% training steps, and a total batch size of 512. The model was trained on the Czech corpus (see Section Training Data) for 2 epochs, or 250,000 steps.

\subsection{Multilingual Distillation}
In this experiment, we trained two models using multilingual distillation \cite{reimers2020making} with the English \textit{all-mpnet-base-v2} model as the teacher, chosen for its high MTEB benchmark \cite{muennighoff2023mteb} performance relative to its size. Despite the original paper using a pre-trained multilingual model to initialize the student model, we found a newly initialized BERT model with a merged Czech-English vocabulary to be more effective. This approach surpassed the performance of pre-trained multilingual models in our experiments.

We addressed the discrepancy in embedding size between the teacher (768) and the student (256) by adding a linear projection to the student model and normalizing all embeddings during training. After training, the projection was removed, resulting in a bilingual student model with the desired embedding size.

We utilized two parallel datasets for our training: a high-quality dataset for non-commercial use, the \textit{czeng20-csmono}~\cite{kocmi2020announcing} and a commercial one, the \textit{Paracrawler v9}~\cite{banon-etal-2020-paracrawl} (see Section Training Data). Our method was efficient, providing successful training on both, which resulted in the \Mse and \CommercialMse models, respectively.

\subsection{Unsupervised Fine-Tuning}

For enhanced performance, all pre-trained language models, including \RetroMAE and those employing multilingual distillation, underwent additional unsupervised fine-tuning. This process aimed to optimize their representations.

In our research, we found that models trained using conventional masked language modelling (MLM) objectives such as BERT, RoBERTa, or even more sophisticated ELECTRA pre-training, often performed sub-optimally in a bi-encoder setup. Therefore, we also fine-tuned \textit{Small-E-Czech} \cite{kocian2022siamese}, a Czech \Small ELECTRA model ~\cite{clark2020electra} , to evaluate the impact of unsupervised fine-tuning on models not tailored for sentence embedding.

\subsubsection{Contrastive learning}

The contrastive learning approaches used Wikipedia data for their experiments. The experiments involved three methodologies: SimCSE \cite{gao2021simcse}, RankCSE \cite{liu2023rankcse}, and InfoCSE \cite{wu2022infocse}. Due to time constraints, we did not conduct an exhaustive hyperparameter grid search. Instead, we consistently adopted the recommended values provided by the original authors of each specific method. In each approach, consistent with the original SimCSE methodology, identical sentences were used as positive pair examples. The sole difference came from a dropout mask applied during creation, leading to minimal augmentation. In-batch negatives were also included.

The AdamW optimizer was used for all experiments, with learning rate 3e-5 for SimCSE and RankSCE, and learning rate 7e-6 for InfoCSE. Each model was trained for 3 epochs with a batch size of 128.

In RankCSE, the SimCSE model variants were utilized as teachers for ranking distillation. In line with \citet{liu2023rankcse}, both ListNet and ListMLE were experimented with, but no significant differences were observed, leading to the adoption of ListNet for further experiments.

\subsubsection{TSDAE}

The TSDAE \cite{wang2021tsdae} is a notable approach used for improving model representations or domain adaptation. However, its training is more computationally and memory intensive than previous methods. TSDAE employs an auto-encoder training, which is considerably slower than RetroMAE as it uses a deep (multi-layer) decoder rather than a shallow one.

In this experiment, Small-E-Czech was fine-tuned with TSDAE using a batch size of 16 and a learning rate of 5e-5 on a single GPU, following the original author's code. The encoder and decoder weights were tied during the training process. The model was trained on a sample of sentences from the Czech corpus for one epoch.

We also attempted to pre-train a BERT model from scratch using the TSDAE method. However, this resulted in exceedingly poor performance on downstream tasks. The trained embeddings might be overly complex for general NLP tasks, requiring a sophisticated decoder, which isn't suitable for typical fine-tuning scenarios that employ a simpler classifier or regressor.

\section{Training Data}
\label{sec:training_data}

The pre-training process utilized a non-public corpus internally called \textit{Czech corpus}, which is 253 GB in size and was obtained by Seznam.cz. This corpus comprises post-processed texts extracted from Czech web pages, encompassing diverse quality levels and lacking a specific domain specification. During the corpus cleaning phase, documents that were deemed too short, non-Czech, duplicated, or classified as spam were excluded.

For the multilingual distillation process, the English-to-Czech transfer was conducted using two distinct datasets. The first one, czeng20-csmono \cite{kocmi2020announcing}, is a high-quality corpus containing 50 million pairs of Czech sentences and their corresponding synthetic English translations, intended strictly for non-commercial use. The second dataset, Paracrawler v9 \cite{banon-etal-2020-paracrawl}, is an open-source resource that also comprises 50 million Czech-English sentence pairs. Despite its less refined nature and origin through bitext mining, the dataset allows commercial applications.

In the unsupervised fine-tuning experiments, the training set was the Czech Wikipedia dump.\footnote{https://dumps.wikimedia.org} Pre-processing steps involved splitting paragraphs into sentences, filtering out special characters and removing sentences deemed too short or too long \cite{gao2021simcse}. Contrastive learning methods, such as SimCSE, RankCSE, InfoCSE, typically utilize a sample size of $10^6$ sentences from the English Wikipedia with appropriate filtering. In this study, a similar approach was extended to include also the full training split from the Czech Wikipedia dataset, resulting in a total of $5.2 \times 10^6$ sentences after pre-processing. All models were fine-tuned using SimCSE on both datasets. Our experiments showed that using the larger dataset, on average, improved the score on the STS task by 1 percentage point for small models and 2 percentage points for base models.

\section{Evaluation}
To explore the performance of each model and understand its behavior across various NLP tasks, multiple evaluation tasks were conducted. Both intrinsic and extrinsic evaluations were performed to thoroughly assess the models.

Intrinsic evaluation aims to test how effectively the embeddings capture the semantic meaning and syntactic structure of sentences. Extrinsic evaluation focuses on the models' performance in a range of NLP challenges within real-world applications and downstream tasks. 

To facilitate these evaluations, publicly available Czech datasets were utilized. In the spirit of transparency, our evaluation pipeline is accessible via the following URL, which will be included in the camera-ready version of the paper.

\subsection{Intrinsic Evaluation}

\subsubsection{Costra}
Costra \citep{barancikova2020} is a dataset designated for evaluating the quality of sentence embeddings spaces. It examines the proficiency of sentence embeddings in capturing intricate linguistic phenomena such as paraphrases, tense, or style. For instance, given a triplet of sentences --  an original, a paraphrased and an antonymous sentence -- the vector similarity (cosine) between the original and paraphrased sentence is expected to be greater than the similarity between the original and antonymous sentence. The percentage of such cases is given as accuracy score.\footnote{The method of evaluation used in our study slightly differs from the one originally proposed by \citet{barancikova2020}. The reason for this difference is straightforward. In their study, \citet{barancikova2020} assess models across six categories: \textit{basic}, \textit{modality}, \textit{time}, \textit{style}, \textit{generalization}, and \textit{opposite}.
However, upon conducting our evaluations, we find that the first two categories (\textit{basic} and \textit{modality}) are too hard for all current models. In these categories, all examined embeddings performed below the random baseline. Consequently, it is unfeasable to distinguish whether good performance in these categories can be attributed to model quality or randomness. In order to keep the evaluation reliable, the \textit{basic} and \textit{modality} categories are omitted from our evaluation and instead only the average performance across the remaining four categories (\textit{time}, \textit{style}, \textit{generalization}, and \textit{opposite}) is reported as our Costra score.
By adjusting the evaluation approach in this manner, we aimed to provide a more accurate and meaningful assessment of the model's performance in the selected categories, while acknowledging the limitations and challenges posed by certain linguistic phenomena in the evaluation process.}

\subsubsection{STS}
In the Semantical Textual Similarity (STS) task, we employ three datasets. The first, CNA, sourced from Sido (2021), contains 1,100 sentence pairs from Czech journalistic texts, using only the test data. The label for each sentence pair was determined by taking the average from 9 different annotations. The second, SVOB-IMG, comprises 850 pairs from image descriptions, and the third, SVOB-HL, has 525 pairs from headlines. Both were translated from the English SentEval dataset \cite{conneau2018senteval} and subsequently annotated \cite{svoboda2018czech}.

\begin{table*}[h!]
    \centering\footnotesize
    \begin{tabular}{l|cccc|c|c}
    \toprule
    & \multicolumn{4}{c}{Spearman's correlation} & \multicolumn{1}{c}{Accuracy} & \multicolumn{1}{c}{P@10}\\
    \textbf{Model} & \textbf{SVOB-IMG} & \textbf{SVOB-HL} & \textbf{CNA} & \textbf{STS-Average} & \textbf{Costra} & \textbf{DaReCzech} \\
    \midrule
    Random baseline & 2.40 & 3.85 & 29.09 & 11.78 & 49.54 & 38.10 $\pm$ 0.31\\
    Random-small & 64.61 & 55.30 & 67.69 & 62.53 & 68.73 & 40.38 $\pm$ 0.35 \\
    Avg. fastText & 54.81 & 47.52 & 72.47 & 58.26 & 65.75 & 37.88 $\pm$ 0.31 \\\midrule
    Small-E-Czech & 39.67 & 42.43 & 62.80 & 48.30 & 64.29 & 37.31 $\pm$ 0.33 \\
    \RetroMAE & 78.88 & 66.21 & 83.82 & 76.30 & 69.66 & 42.16 $\pm$ 0.36 \\
    \CommercialMse & 90.11 & 77.66 & 84.99 & 84.25 & 70.42 & 42.33 $\pm$ 0.32 \\
    \Mse & \textbf{90.94} & 83.89 & 87.97 & 87.60 & 71.22 & 42.01 $\pm$ 0.37 \\\midrule
    SimCSE-Small-E-Czech & 61.70 & 59.75 & 77.26 & 66.24 & 66.44 &  39.20  $\pm$ 0.38 \\
    TSDAE-Small-E-Czech & 77.45 & 66.17 & 83.16 & 75.59 & 69.42 & 40.54 $\pm$ 0.37 \\
    SimCSE-RetroMAE & 78.88 & 71.92 & 85.19 & 78.66 & 69.63 &  42.04 $\pm$ 0.37 \\
    RankCSE-RetroMAE & 79.91 & 72.03 & 85.10 & 79.01 & 69.79 & 41.97 $\pm$ 0.36 \\
    InfoCSE-RetroMAE & 79.30 & 65.58 & 84.31 & 76.40 & 69.89 & 41.77 $\pm$ 0.38 \\

    SimCSE-\CommercialMse & 90.29 & 78.80 & 85.91 & 85.00 & 71.12 & \textbf{42.38} $\pm$ 0.35 \\
    SimCSE-\Mse & 90.73 & \textbf{84.22} & \textbf{88.56} & \textbf{87.83} & \textbf{71.77} & 42.18 $\pm$ 0.38\\\midrule
    OpenAI Ada Embedding& 83.51 & 78.04 & 86.21 & 82.59 & 69.01 & 42.21 $\pm$ 0.31 \\
    \bottomrule
    \end{tabular}
    \caption{Zero-shot evaluation of our models and several baselines on STS, Costra and DaReCzech. The table is organized into four horizontal sections: baselines, pre-trained models, models fine-tuned for sentence embeddings, and external sentence embedding services.}
    \label{table:results_zero_shot}
\end{table*}

\subsection{Extrinsic Evaluation}
\subsubsection{Multi-label classification}
The Czech text document corpus CTDC \cite{kral2017czech} is a dataset designed for direct comparison of document classification on Czech data. It comprises 12,000 news articles labeled with 37 categories used for classification. A \textit{5-fold cross-validation} procedure was conducted. A micro-averaged F1 score was used as an evaluation metric, reported with a standard deviation.
\subsubsection{Sentiment analysis}
The Czech Facebook Dataset (CFD) \cite{habernal2013sentiment} was employed for sentiment analysis. It consists of 2,587 positive, 5,174 neutral, and 1,991 negative posts. To assess model performance, a \textit{10-fold cross-validation} procedure was conducted, following the methodology outlined in~\citet{straka2021robeczech}. The evaluation metric used was the macro-averaged F1 score, and the reported results include the standard deviation.
\subsubsection{Relevance ranking}
Ranking documents based on user queries is a fundamental task in information retrieval, with applications in search engines and recommendation systems. To evaluate model performance in this context, the DaReCzech dataset \cite{kocian2022siamese} was utilized. This dataset consists of 1.6 million Czech query-document pairs split into training, development and testing sets, each labeled with a relevance score.

In assessing the effectiveness of ranking algorithms, the precision at 10 (\textit{p@10}) metric was employed. This metric is commonly used in information retrieval to measure how well the model performs in returning the relevant documents within the top 10 results for a given query.

\section{Experiments}
We present an evaluation of semantic models' quality in various settings, including zero-shot evaluation, linear-probing experiments, and whole model fine-tuning.

\subsection{Zero-shot Evaluation}

Zero-shot evaluation assesses a model's generalization capabilities through embedding space quality measured with Costra and STS. Additionally, we conduct a zero-shot evaluation on DaReCzech, based on the concept that relevant documents typically show high semantic similarity to a query.

We use cosine similarity and CLS pooling~\cite{devlin2018bert}. However, for models that have been pre-trained using methods such as MLM, the choice of pooling method can greatly influence their performance in zero-shot scenarios. In these cases, MEAN or MAX pooling methods tend to give the best results. We determine the optimal setting by using the Spearman's correlation with the CNA train dataset. In the case of Small-E-Czech the MEAN pooling is used.

Alongside our trained models, we evaluate a sentence embedding service for comparison, specifically the OpenAI's Ada embeddings (\textit{text-embedding-ada-002}). These multilingual embeddings, each with a length of 1536, %are versatile for various NLP tasks. They 
are expected to provide quality semantic features and a robust embedding space, making them suitable for zero-shot settings. In measuring similarity, we adhere to official recommendation of using cosine similarity. Although these embeddings could be a viable solution for industrial applications, their cost-effectiveness may be questionable for larger datasets, rendering them potentially unsuitable.

For our baselines, we employ the \textit{random baseline} with random embeddings, \textit{random-small} which is a model with the \Small ELECTRA architecture initialized randomly without pre-training, and averaged fastText \cite{bojanowski2017enriching} embeddings. We use the MEAN pooling for the random-small.

\subsubsection{Zero-shot results}
As shown in Table \ref{table:results_zero_shot}, the initial performance of Small-E-Czech's embeddings in zero-shot evaluations was poor, but improved significantly with unsupervised methods such as SimCSE and TSDAE. Despite InfoCSE and RankCSE being theoretically advanced successors of SimCSE, they did not exhibit any improvements over their predecessor in our experiments. Therefore, in the remainder of this work, we will only present the performance of SimCSE.

RetroMAE pre-training yielded even better results than the SimCSE-Small-E-Czech. Multilingual distillation performed exceptionally, even surpassing OpenAI embeddings on all datasets despite having 6x smaller embeddings.

RetroMAE and distilled models do not exhibit significant enhancements in STS after undergoing SimCSE training. The lack of improvement is attributed to the already favorable spatial properties of the embeddings generated by these models. %, posing a challenge for further enhancement. 
This pattern is obvious in distilled models, which imitate a model trained with a contrastive objective but using supervised data. However, the situation is not immediately clear for autoencoder models like TSDAE and RetroMAE, as they display similar behavior. This suggests that autoencoders and contrastive learning may have some degree of interchangeability.

\subsection{Linear Probing}

\begin{table}[!htb]
    \centering\footnotesize
    \begin{tabular}{lll}\toprule
    & \multicolumn{2}{c}{F1 score}\\
    \textbf{Model} & \multicolumn{1}{c}{\textbf{CFD}} & \multicolumn{1}{c}{\textbf{CTDC}} \\\midrule
    Random baseline & 23.11 $\pm$ 0.49 & 21.44 $\pm$ 0.24 \\
    Random-small & 25.66 $\pm$ 1.21 & 21.45 $\pm$ 0.23 \\
    Avg. fastText & 64.12 $\pm$ 1.41 & 66.56 $\pm$ 0.58 \\\midrule
    Small-E-Czech & 32.38 $\pm$ 2.35 & 25.92 $\pm$ 0.56 \\
    RetroMAE-small & 68.56 $\pm$ 1.37 & 78.18 $\pm$ 0.08 \\
    \CommercialMse & 71.30 $\pm$ 1.40 & 80.18 $\pm$ 0.10 \\
    \Mse & 72.84 $\pm$ 1.62 & \textbf{82.38} $\pm$ 0.25 \\\midrule
    
    SimCSE-Small-E-Czech & 54.78 $\pm$ 2.59 & 50.06 $\pm$ 0.45 \\
    SimCSE-RetroMAE-small & 68.70 $\pm$ 1.49 & 77.32 $\pm$ 0.39 \\
    TSDAE-Small-E-Czech & 66.99 $\pm$ 1.24 & 75.84 $\pm$ 0.24\\
    SimCSE-\CommercialMse & 71.87 $\pm$ 1.06 & 79.41 $\pm$ 0.28 \\
    SimCSE-\Mse & 72.66 $\pm$ 1.35 & 81.63 $\pm$ 0.37 \\
    \midrule
    
    OpenAI Ada Embedding & \textbf{75.20} $\pm$ 1.12 & 75.89 $\pm$ 0.46 \\\midrule    
    
    \end{tabular}
    \caption{Linear probing evaluation of our models on CFD and CTDC.}
    \label{table:results_linear_probing}
\end{table}

Linear probing clarifies a model's internal representations, which are not explicitly discernible based on a position in a vector space, and provides a novel perspective on sentence embeddings, making them suitable for tasks where cosine similarity is inadequate. We train a classifier head on top of pre-trained sentence embeddings and evaluate its performance on CFD and CTDC,
reserving other datasets for the sentence-pair bi-encoder setup using simple cosine similarity.
We chose CLS pooling as the standard for all models due to the classification head's adaptability, which reduces disparities between different pooling methods in our tests.

\subsubsection{Linear probing results}
As outlined in Table \ref{table:results_linear_probing}, ELECTRA pre-training enhances the performance of linear probing compared to a randomly initialized model, yet it remains less effective than other methods.
This may be attributed to the anisotropy problem observed within the semantic spaces of Small-E-Czech embeddings. In other words, the cosine similarity of nearly any meaningful texts is very close to 1, limiting their expressiveness. While the embeddings can be effectively redistributed through full model fine-tuning, the direct utilization of embeddings may prove inadequate, leading to subpar performance.

Contrastive methods, designed to enhance the spatial properties of sentence embeddings, notably also enrich the generated features. This effect helps Small-E-Czech to reduce the performance gap, leading to significant score improvements of about 22 and 24 percentage points on the CFD and CTDC datasets, respectively. However, it still doesn't surpass fastText, which performs notably better in linear-probing than in zero-shot. TSDAE furthers these gains, exceeding fastText. However, such improvements are not observed in our other models after contrastive fine-tuning, where there is insignificant change in performance. \RetroMAE consistently generates robust features for linear probing, and bilingual models consistently outperform others,
particularly on the CTDC dataset in a linear-probing setup.

OpenAI's embeddings top the sentiment analysis rankings, possibly benefiting from the abundance of data and the task's prevalence. However, for more complex tasks like CTDC, our models outperform them.

\subsection{Fine-tuning}

\begin{table*}[!htb]
    \centering\footnotesize
    \begin{tabular}{llll}\toprule
    & \multicolumn{2}{c}{F1 score} & \multicolumn{1}{c}{P@10}\\
    \textbf{Model} & \multicolumn{1}{c}{\textbf{CFD}} & \multicolumn{1}{c}{\textbf{CTDC}} & \multicolumn{1}{c}{\textbf{DaReCzech}}\\\midrule
    Random-small & 69.67 $\pm$ 1.35 & 40.32 $\pm$ 1.40 & 42.66 $\pm$ 0.35 \\\midrule
    
    Small-E-Czech & 76.94 $\pm$ 1.18 & 58.12 $\pm$ 1.52 & 43.64 $\pm$ 0.37 \\
    \RetroMAE & 76.85 $\pm$ 1.16 & 84.58 $\pm$ 0.37 & 45.29 $\pm$ 0.34 \\
    \CommercialMse & 77.42 $\pm$ 1.60 & 86.02 $\pm$ 0.12 & 45.55 $\pm$ 0.33 \\
    \Mse & \textbf{78.73} $\pm$ 1.39 & 85.85 $\pm$ 0.21 & \textbf{45.75} $\pm$ 0.34 \\
    \midrule

    SimCSE-Small-E-Czech & 76.27 $\pm$ 1.19 & 68.33 $\pm$ 1.34 & 44.64 $\pm$ 0.38 \\
    SimCSE-RetroMAE & 76.16 $\pm$ 1.53 & 84.95 $\pm$ 0.05 & 45.26 $\pm$ 0.34 \\
    TSDAE-Small-E-Czech & 75.31 $\pm$ 1.00 & 73.15 $\pm$ 0.29 & 44.84 $\pm$ 0.35 \\

    SimCSE-\CommercialMse & 77.31 $\pm$ 1.40 & \textbf{86.10} $\pm$ 0.32 & 45.66 $\pm$ 0.33 \\
    SimCSE-\Mse & \textbf{78.73} $\pm$ 1.43 & 85.25 $\pm$ 1.15 & \textbf{45.75} $\pm$ 0.33 \\
    \midrule
    
    \end{tabular}
    \caption{Fine-tuning evaluation of our models on CFD, CTDC and DaReCzech datasets.}
    \label{table:results_finetune}
\end{table*}

In this experiment, we permit models to directly optimize their sentence embeddings for specific tasks, enabling an in-depth assessment of their potential on the DaReCzech, CFD, and CTDC datasets. We continue optimizing the classification head, as we previously did with CFD and CTDC. However, for the DaReCzech training, we exclude a classification head, maintaining it within a bi-encoder setup, with the optimization focused solely on the embedding space to closely align a query with its relevant documents.

We do not fine-tune OpenAI's embeddings as we lack access to the model and control over the fine-tuning process.

\subsubsection{Fine-tuning results}
The results in Table \ref{table:results_finetune} reveal that ELECTRA pre-training effectively adapts its sentence embeddings to the given task, achieving commendable performance, confirming the results of \citet{kocian2022siamese}. Despite contrastive learning enhancing fine-tuning performance for tasks involving sentence pair comparisons, like DaReCzech, it can lead to a decline in performance for other tasks, which confirms previous findings of \citet{gao2021simcse}. However, this isn't the case for Small-E-Czech augmented with SimCSE and subsequently fine-tuned on the CTDC dataset, where it improves by another 10 points. This suggests that contrastive learning can still be advantageous for fine-tuning specific types of downstream tasks.

RetroMAE's pre-training demonstrates strong performance in these settings, showing its capacity to learn powerful encodings for any piece of text. Notably, the fine-tuned bilingual models surpass others in this setup, making them a generally excellent choice when both a robust English model and a large bilingual dataset are available.

\section{Other Experiments}

\subsection{Comparison to existing Czech \Base models}

\begin{table*}[h!]
    \centering\footnotesize
    \begin{tabular}{lcccccc}
    \toprule
    \textbf{Model} & \textbf{Costra} & \textbf{STS-Average} & \textbf{CFD} & \textbf{DaReCzech} & \textbf{CTDC} \\
    \midrule
    Czert-b-base-cased   & \textbf{72.08}  & 74.79  & 78.73 $\pm$ 1.25  & 45.63 $\pm$ 0.34  & 88.69 $\pm$ 0.21    \\
    FERNET-C5     & 67.57    & 65.46  & \textbf{82.00} $\pm$ 1.43         & 45.87 $\pm$ 0.34  & \textbf{89.56} $\pm$ 0.19    \\
    RobeCzech     & 63.94    & 69.41 & 80.54 $\pm$ 0.93 & 45.54 $\pm$ 0.31 & 86.01 $\pm$ 0.24  \\
    LaBSE         & 70.63    & 82.91 & 79.79 $\pm$ 1.07  & \textbf{46.15} $\pm$ 0.34  & 87.97 $\pm$ 0.31    \\\midrule
    \Mse          & 71.22    & \textbf{87.60}  & 78.73 $\pm$ 1.39  & 45.75 +\textbf{}-0.34  & 85.85 $\pm$ 0.21    \\
    \CommercialMse      & 70.42    & 84.25  & 77.42 $\pm$ 1.60  & 45.55 $\pm$ 0.33  & 86.02 $\pm$ 0.12 \\    
    \bottomrule
    \end{tabular}
    \caption{Comparison of \Base Czech models (top four) to two of our best \Small models (bottom -- \Mse, \CommercialMse) on all datasets.}
    \label{table:base_vs_small}
\end{table*}

We compared the performance of our distilled models (\CommercialMse, \Mse) with the existing Czech \Base models: \textit{Czert-b-base-cased}~\cite{sido2021czert}, \textit{FERNET-C5}~\cite{lehevcka2021comparison}, \textit{RobeCzech}~\cite{straka2021robeczech} and multilingual \textit{LaBSE}~\cite{feng2022language}. As shown in Table~\ref{table:base_vs_small}, despite being approximately 8 times smaller than the existing Czech \Base-sized model, our \Small-sized models have proven to be highly competitive. The intrinsic evaluation of their embeddings' semantic features (\textit{STS-AVERAGE}, \textit{COSTRA}) revealed that they outperform the majority of BASE-sized systems. Surprisingly, our models perform on par with most BASE-sized models in relevance ranking (DaReCzech) and are only surpassed by the LABSE model, whose embeddings have a semantic origin.

\subsection{Fine-tuning data size}

\begin{figure}[!htb]
    \centering
    \includegraphics[scale=0.53]{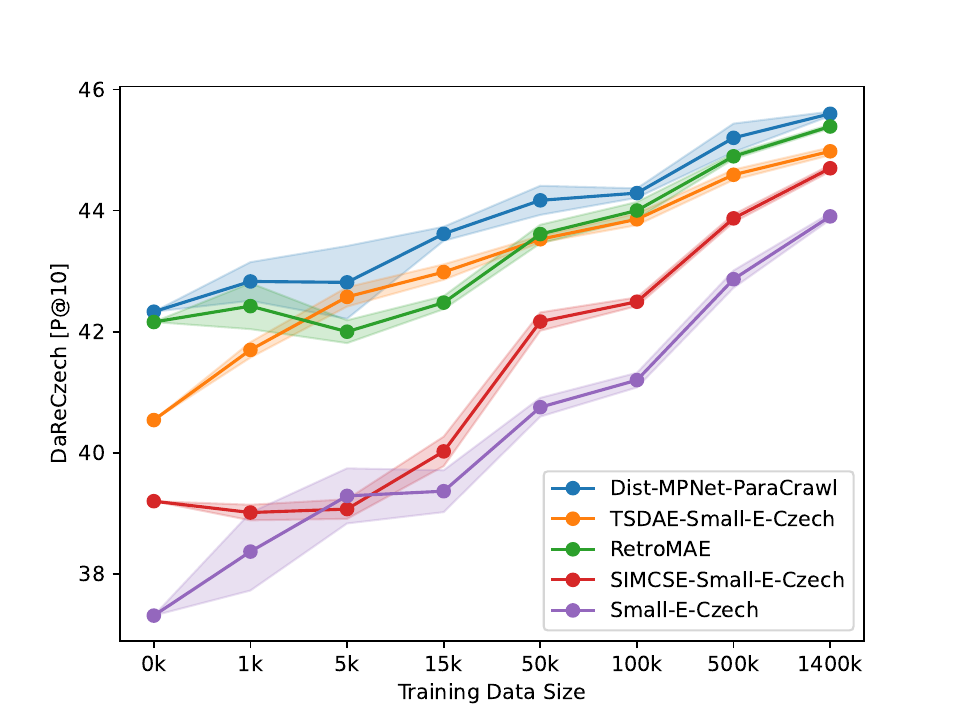}
    \caption{Comparison of different models when fine-tuned with various training data sizes evaluated on the DaReCzech test set. Each data point represents the average P@10 of four models, with standard deviation indicated.}
    \label{fig:finetuning_datasize}
\end{figure}

One of our hypotheses posited that ample data for downstream tasks during model fine-tuning could ameliorate differences among initial models. To empirically validate this hypothesis, we leveraged the DaReCzech training dataset, comprising in excess of 1.4 million query-document pairs, and proceeded to train diverse models on its subsets. Specifically, each model underwent training across four instances, each involving random subsets of sizes 1\,000; 5\,000; 10\,000; 100\,000; 500\,000; and 1\,400\,000. The aggregate P@10 values, alongside their corresponding standard deviations, are visualised in Figure~\ref{fig:finetuning_datasize}. Evidently, the distinctions among initial models remains discernible despite variations in data size, even following comprehensive fine-tuning with the entire available dataset. 

\section{Model Usage in Seznam.cz}

Seznam.cz started its proprietary search engine in 2005, initially solely based on lexical matching. In 2021, a semantic branch was integrated, utilizing bi-encoder neural networks, notably Small-E-Czech model. This augmentation extended the search engine's capabilities beyond mere word matching, enabling deeper semantic comprehension of user queries and web content. Consequently, search accuracy and relevance has been significantly improved, providing users with more precise search results. Small-E-Czech has also proven valuable in additional tasks at Seznam.cz, such as query correction~\cite{filip_rychla_Year}, named-entity recognition~\cite{havova_jak} or clickbait article detection. 

Recognizing the limitations of the Small-E-Czech model, as previously discussed, we have gradually replaced it in our primarily retrieval-based applications with its semantic alternatives, namely the \RetroMAE and \CommercialMse models. These models now hold a crucial role, particularly in organic search, featured snippets, and image search, enhancing the overall search experience. % for our users.

\subsection{Organic Search}

Seznam.cz's search engine utilizes a sequence of cascade stages for organic search, each characterized by varying complexity and time constraints \cite{kocian2022siamese}. The initial stage operates across the entire document index, demanding rapid processing. Subsequent stages handle documents pre-selected by prior stages, they are allowed longer processing times to retrieve only documents relevant to user queries. Across all these stages, the Small-E-Czech model has consistently maintained a crucial role.

In the final component of the organic search (\textit{stage-2} Catboost~\cite{prokhorenkova2018catboost} tree), Small-E-Czech-based bi-encoder model trained on the DaReCzech dataset has been employed as an additional feature leading to 1.5 percentage points improvement on the offline test set and a 3.8 percentage points improvement, as assessed by human annotators after deployment in production (measured using P@10)~\cite{kocian2022siamese}. Presently, this model has been substituted with the new \RetroMAE model, yielding additional 2 percentage points improvement. This change has led to an improved user experience with our search functionality. Upon analysis, %it becomes evident that 
the model primarily enhances results for intricate and information-rich queries.

We are currently engaged in an active effort to replace outdated models in the preceding stages.

\subsection{Featured Snippets}

Since 2022, Seznam.cz introduced featured snippets to its search results, providing direct answers from documents on the search engine results page (SERP) alongside organic results. Two-step architecture is employed: a bi-encoder first pre-selects relevant paragraphs, and a cross-encoder subsequently re-ranks them. In this process, the Small-E-Czech model, enhanced with annotations, plays a key role.

Unlike organic results, featured snippets are considered an optional addition. Hence, it is essential to display only accurate and reliable featured snippets. To ensure this, we establish a threshold, based on our annotated data, ensuring at least 95\% accuracy for the snippets. Accordingly, our primary metric is recall, conditioned on the model maintaining a minimum precision of 95\%.

Despite the already high recall of our retrieval model, there are instances when the bi-encoder falls short in identifying the appropriate paragraph. Transitioning from the Small-E-Czech model to \RetroMAE reduced this error rate by a significant 20\%. Consequently, our refined pipeline has resulted in an average 3.5\% boost in recall, with a consistent precision of 95\%.

\subsection{Image Search}

To accommodate the image queries in Seznam.cz,
the~engine employs a joint embedding space for both images and texts. Texts associated with specific images form cohesive clusters in the embedding space, while texts and images lacking semantic relevance remain farther apart. This embedding space is developed through joint training of both text and image encoders, where the model learns to understand the connections between texts and images much like \textit{CLIP}~\cite{radford2021learning}.

Our ultimate metric for evaluating the quality of the embedding space is the performance of the production image search model, which uses the embedding space similarity of query and image as one of its features. By replacing the Small-E-Czech text encoder with the \CommercialMse model, the relative NDCG (normalized discounted cumulative gain) error rate reduced by 3.2\%, similar to the improvement observed when the Small-E-Czech text encoder replaced the original FastText encoder. Notably, this improvement is significant considering that the production model comprises numerous complex features and has limited room for enhancement.

We also assess the quality of the embedding space separately and compare different text encoders. In this comparison, \CommercialMse allows retrieving 7\% more relevant images than the previous best Small-E-Czech model.

\subsection{Deployment Information}

In our deployment configuration, both organic and featured snippet's components are computed on the CPU with AVX512 instructions.
Our models are stored in the ONNX format and quantized using FP16, with negligible impact on performance. Processing efficiency is maintained through our in-house queuing system, integrated with Kubernetes. Documents and paragraphs are processed offline whenever there is a modification or a new document is identified, while queries are handled in real-time. The average query processing time is approximately 4 milliseconds. Regarding images, similar to queries, computations are conducted online only for queries, and images are embedded using image-specific models developed in-house.

\section{Conclusion}
In this work, we trained and evaluated multiple \Small-sized Czech models for sentence embeddings. Despite their relative small size, these models matched or exceeded the performance of \Base models across diverse tasks. The intrinsic evaluations proved the models' effectiveness in capturing semantic and syntactic information, while the extrinsic tasks demonstrated their practical use in real-world applications in NLP. We also established that a powerful bilingual model can be derived through multilingual distillation from a proficient English model, outperforming several pre-trained Czech models. The effectiveness of RetroMAE's pre-training and language distillation methods was confirmed in various Seznam.cz tasks. We anticipate our released models could promote wider applications and inspire further exploration into efficient, smaller-scale models for complex language tasks.

\section{Acknowledgments}

We thank Barbora Rišová, Martin Dvořák, Martin Habrovec and Dominika Kozlová for experiments done on image-text encoders.
We also thank Milan Straka and Jana Straková for their valuable comments on this article.

% AAAI is especially grateful to Peter Patel Schneider for his work in implementing the original aaai.sty file, liberally using the ideas of other style hackers, including Barbara Beeton. We also acknowledge with thanks the work of George Ferguson for his guide to using the style and BibTeX files --- which has been incorporated into this document --- and Hans Guesgen, who provided several timely modifications, as well as the many others who have, from time to time, sent in suggestions on improvements to the AAAI style. We are especially grateful to Francisco Cruz, Marc Pujol-Gonzalez, and Mico Loretan for the improvements to the Bib\TeX{} and \LaTeX{} files made in 2020.

% The preparation of the \LaTeX{} and Bib\TeX{} files that implement these instructions was supported by Schlumberger Palo Alto Research, AT\&T Bell Laboratories, Morgan Kaufmann Publishers, The Live Oak Press, LLC, and AAAI Press. Bibliography style changes were added by Sunil Issar. \verb+\+pubnote was added by J. Scott Penberthy. George Ferguson added support for printing the AAAI copyright slug. Additional changes to aaai24.sty and aaai24.bst have been made by Francisco Cruz, Marc Pujol-Gonzalez, and Mico Loretan.

% \bigskip
% \noindent Thank you for reading these instructions carefully. We look forward to receiving your electronic files!

\bibliography{aaai24}

\begin{thebibliography}{44}
\providecommand{\natexlab}[1]{#1}

\bibitem[{Ba{\~n}{\'o}n et~al.(2020)Ba{\~n}{\'o}n, Chen, Haddow, Heafield, Hoang, Espl{\`a}-Gomis, Forcada, Kamran, Kirefu, Koehn, Ortiz~Rojas, Pla~Sempere, Ram{\'\i}rez-S{\'a}nchez, Sarr{\'\i}as, Strelec, Thompson, Waites, Wiggins, and Zaragoza}]{banon-etal-2020-paracrawl}
Ba{\~n}{\'o}n, M.; Chen, P.; Haddow, B.; Heafield, K.; Hoang, H.; Espl{\`a}-Gomis, M.; Forcada, M.~L.; Kamran, A.; Kirefu, F.; Koehn, P.; Ortiz~Rojas, S.; Pla~Sempere, L.; Ram{\'\i}rez-S{\'a}nchez, G.; Sarr{\'\i}as, E.; Strelec, M.; Thompson, B.; Waites, W.; Wiggins, D.; and Zaragoza, J. 2020.
\newblock {P}ara{C}rawl: Web-Scale Acquisition of Parallel Corpora.
\newblock In \emph{ACL}.

\bibitem[{Barančíková and Bojar(2020)}]{barancikova2020}
Barančíková, P.; and Bojar, O. 2020.
\newblock Costra 1.1: An Inquiry into Geometric Properties of Sentence Spaces.
\newblock In \emph{TSD}.

\bibitem[{Bojanowski et~al.(2017)Bojanowski, Grave, Joulin, and Mikolov}]{bojanowski2017enriching}
Bojanowski, P.; Grave, E.; Joulin, A.; and Mikolov, T. 2017.
\newblock Enriching word vectors with subword information.
\newblock \emph{TACL}.

\bibitem[{Chuang et~al.(2022)Chuang, Dangovski, Luo, Zhang, Chang, Solja{\v{c}}i{\'c}, Li, Yih, Kim, and Glass}]{chuang2022diffcse}
Chuang, Y.-S.; Dangovski, R.; Luo, H.; Zhang, Y.; Chang, S.; Solja{\v{c}}i{\'c}, M.; Li, S.-W.; Yih, W.-t.; Kim, Y.; and Glass, J. 2022.
\newblock DiffCSE: Difference-based contrastive learning for sentence embeddings.
\newblock In \emph{NAACL}.

\bibitem[{Clark et~al.(2020)Clark, Luong, Le, and Manning}]{clark2020electra}
Clark, K.; Luong, M.-T.; Le, Q.~V.; and Manning, C.~D. 2020.
\newblock {ELECTRA}: Pre-training Text Encoders as Discriminators Rather Than Generators.
\newblock In \emph{ICLR}.

\bibitem[{Conneau and Kiela(2018)}]{conneau2018senteval}
Conneau, A.; and Kiela, D. 2018.
\newblock {S}ent{E}val: An Evaluation Toolkit for Universal Sentence Representations.
\newblock In \emph{LREC}.

\bibitem[{Devlin et~al.(2019)Devlin, Chang, Lee, and Toutanova}]{devlin2018bert}
Devlin, J.; Chang, M.-W.; Lee, K.; and Toutanova, K. 2019.
\newblock {BERT}: Pre-training of Deep Bidirectional Transformers for Language Understanding.
\newblock In \emph{ACL}.

\bibitem[{Feng et~al.(2022)Feng, Yang, Cer, Arivazhagan, and Wang}]{feng2022language}
Feng, F.; Yang, Y.; Cer, D.; Arivazhagan, N.; and Wang, W. 2022.
\newblock Language-agnostic BERT Sentence Embedding.
\newblock In \emph{ACL}.

\bibitem[{Filip(2021)}]{filip_rychla_Year}
Filip, O. 2021.
\newblock Rychlá oprava dotazů ve vyhledávači pomocí neuronových sítí.
\newblock \url{https://www.root.cz/clanky/rychla-oprava-dotazu-ve-vyhledavaci-pomoci-neuronovych-siti/}.

\bibitem[{Gao and Callan(2021)}]{gao2021condenser}
Gao, L.; and Callan, J. 2021.
\newblock Condenser: a Pre-training Architecture for Dense Retrieval.
\newblock In \emph{EMNLP}.

\bibitem[{Gao and Callan(2022)}]{gao2021unsupervised}
Gao, L.; and Callan, J. 2022.
\newblock Unsupervised Corpus Aware Language Model Pre-training for Dense Passage Retrieval.
\newblock In \emph{ACL}.

\bibitem[{Gao, Yao, and Chen(2021)}]{gao2021simcse}
Gao, T.; Yao, X.; and Chen, D. 2021.
\newblock {S}im{CSE}: Simple Contrastive Learning of Sentence Embeddings.
\newblock In \emph{EMNLP}.

\bibitem[{Habernal, Pt{\'a}{\v{c}}ek, and Steinberger(2013)}]{habernal2013sentiment}
Habernal, I.; Pt{\'a}{\v{c}}ek, T.; and Steinberger, J. 2013.
\newblock Sentiment Analysis in {C}zech Social Media Using Supervised Machine Learning.
\newblock In \emph{WASSA}.

\bibitem[{H{\'a}vov{\'a}(2023)}]{havova_jak}
H{\'a}vov{\'a}, M. 2023.
\newblock Jak Vyhledávání na Seznamu rozpozná jm{\'e}na, p{\v{r}}{\'i}jmen{\'i} a osobnosti?
\newblock \url{https://blog.seznam.cz/2023/07/vylepsili-jsme-rozpoznavani-jmen-prijmeni-a-osobnosti-ve-vyhledavani-na-seznamu/}.

\bibitem[{Hinton, Vinyals, and Dean(2015)}]{hinton2015distilling}
Hinton, G.; Vinyals, O.; and Dean, J. 2015.
\newblock Distilling the Knowledge in a Neural Network.
\newblock \emph{arXiv preprint arXiv:1503.02531}.

\bibitem[{Hofst{\"a}tter et~al.(2022)Hofst{\"a}tter, Khattab, Althammer, Sertkan, and Hanbury}]{hofstatter2022introducing}
Hofst{\"a}tter, S.; Khattab, O.; Althammer, S.; Sertkan, M.; and Hanbury, A. 2022.
\newblock Introducing Neural Bag of Whole-Words with ColBERTer: Contextualized Late Interactions using Enhanced Reduction.
\newblock In \emph{CIKM}.

\bibitem[{Khattab and Zaharia(2020)}]{khattab2020colbert}
Khattab, O.; and Zaharia, M. 2020.
\newblock ColBERT: Efficient and Effective Passage Search via Contextualized Late Interaction over BERT.
\newblock In \emph{SIGIR}.

\bibitem[{Kingma and Ba(2014)}]{kingma2014adam}
Kingma, D.; and Ba, J. 2014.
\newblock {Adam: A Method for Stochastic Optimization}.
\newblock \emph{ICLR}.

\bibitem[{Koci{\'a}n et~al.(2022)Koci{\'a}n, N{\'a}plava, {\v{S}}tancl, and Kadlec}]{kocian2022siamese}
Koci{\'a}n, M.; N{\'a}plava, J.; {\v{S}}tancl, D.; and Kadlec, V. 2022.
\newblock Siamese BERT-Based Model for Web Search Relevance Ranking Evaluated on a New Czech Dataset.
\newblock In \emph{AAAI}.

\bibitem[{Kocmi, Popel, and Bojar(2020)}]{kocmi2020announcing}
Kocmi, T.; Popel, M.; and Bojar, O. 2020.
\newblock Announcing CzEng 2.0 Parallel Corpus with over 2 Gigawords.
\newblock \emph{arXiv preprint arXiv:2007.03006}.

\bibitem[{Kong et~al.(2022)Kong, Khadanga, Li, Gupta, Zhang, Xu, and Bendersky}]{kong2022multi}
Kong, W.; Khadanga, S.; Li, C.; Gupta, S.~K.; Zhang, M.; Xu, W.; and Bendersky, M. 2022.
\newblock Multi-Aspect Dense Retrieval.
\newblock In \emph{SIGKDD}.

\bibitem[{Kral and Lenc(2018)}]{kral2017czech}
Kral, P.; and Lenc, L. 2018.
\newblock Czech Text Document Corpus v 2.0.
\newblock In \emph{LREC}.

\bibitem[{Lehe{\v{c}}ka and {\v{S}}vec(2021)}]{lehevcka2021comparison}
Lehe{\v{c}}ka, J.; and {\v{S}}vec, J. 2021.
\newblock Comparison of Czech Transformers on Text Classification Tasks.
\newblock In \emph{SLPS}. Springer.

\bibitem[{Liu et~al.(2021)Liu, Jiao, Massiah, Yilmaz, and Havrylov}]{liu2021trans}
Liu, F.; Jiao, Y.; Massiah, J.; Yilmaz, E.; and Havrylov, S. 2021.
\newblock Trans-encoder: unsupervised sentence-pair modelling through self-and mutual-distillations.
\newblock \emph{arXiv preprint arXiv:2109.13059}.

\bibitem[{Liu et~al.(2023)Liu, Liu, Wang, Wang, Wu, Xian, Zhao, Chen, and Yan}]{liu2023rankcse}
Liu, J.; Liu, J.; Wang, Q.; Wang, J.; Wu, W.; Xian, Y.; Zhao, D.; Chen, K.; and Yan, R. 2023.
\newblock {R}ank{CSE}: Unsupervised Sentence Representations Learning via Learning to Rank.
\newblock In \emph{ACL}.

\bibitem[{Loshchilov and Hutter(2017)}]{loshchilov2017sgdr}
Loshchilov, I.; and Hutter, F. 2017.
\newblock {SGDR}: Stochastic Gradient Descent with Warm Restarts.
\newblock In \emph{ICLR}.

\bibitem[{Loshchilov and Hutter(2019)}]{loshchilov2018decoupled}
Loshchilov, I.; and Hutter, F. 2019.
\newblock Decoupled Weight Decay Regularization.
\newblock In \emph{ICLR}.

\bibitem[{Lu et~al.(2021)Lu, He, Xiong, Ke, Malik, Dou, Bennett, Liu, and Overwijk}]{lu2021less}
Lu, S.; He, D.; Xiong, C.; Ke, G.; Malik, W.; Dou, Z.; Bennett, P.; Liu, T.-Y.; and Overwijk, A. 2021.
\newblock Less is More: Pretrain a Strong Siamese Encoder for Dense Text Retrieval Using a Weak Decoder.
\newblock In \emph{EMNLP}.

\bibitem[{Muennighoff et~al.(2023)Muennighoff, Tazi, Magne, and Reimers}]{muennighoff2023mteb}
Muennighoff, N.; Tazi, N.; Magne, L.; and Reimers, N. 2023.
\newblock MTEB: Massive Text Embedding Benchmark.
\newblock arXiv:2210.07316.

\bibitem[{Prokhorenkova et~al.(2018)Prokhorenkova, Gusev, Vorobev, Dorogush, and Gulin}]{prokhorenkova2018catboost}
Prokhorenkova, L.; Gusev, G.; Vorobev, A.; Dorogush, A.~V.; and Gulin, A. 2018.
\newblock CatBoost: unbiased boosting with categorical features.
\newblock \emph{NeurIPS}, 31.

\bibitem[{Qu et~al.(2021)Qu, Ding, Liu, Liu, Ren, Zhao, Dong, Wu, and Wang}]{qu2020rocketqa}
Qu, Y.; Ding, Y.; Liu, J.; Liu, K.; Ren, R.; Zhao, W.~X.; Dong, D.; Wu, H.; and Wang, H. 2021.
\newblock {R}ocket{QA}: An Optimized Training Approach to Dense Passage Retrieval for Open-Domain Question Answering.
\newblock In \emph{NAACL}.

\bibitem[{Radford et~al.(2021)Radford, Kim, Hallacy, Ramesh, Goh, Agarwal, Sastry, Askell, Mishkin, Clark et~al.}]{radford2021learning}
Radford, A.; Kim, J.~W.; Hallacy, C.; Ramesh, A.; Goh, G.; Agarwal, S.; Sastry, G.; Askell, A.; Mishkin, P.; Clark, J.; et~al. 2021.
\newblock Learning transferable visual models from natural language supervision.
\newblock In \emph{ICML}, 8748--8763.

\bibitem[{Reimers and Gurevych(2019)}]{reimers2019sentence}
Reimers, N.; and Gurevych, I. 2019.
\newblock Sentence-{BERT}: Sentence Embeddings using {S}iamese {BERT}-Networks.
\newblock In \emph{EMNLP}.

\bibitem[{Reimers and Gurevych(2020)}]{reimers2020making}
Reimers, N.; and Gurevych, I. 2020.
\newblock Making Monolingual Sentence Embeddings Multilingual using Knowledge Distillation.
\newblock In \emph{EMNLP}.

\bibitem[{Sido et~al.(2021)Sido, Pra{\v{z}}{\'a}k, P{\v{r}}ib{\'a}{\v{n}}, Pa{\v{s}}ek, Sej{\'a}k, and Konop{\'\i}k}]{sido2021czert}
Sido, J.; Pra{\v{z}}{\'a}k, O.; P{\v{r}}ib{\'a}{\v{n}}, P.; Pa{\v{s}}ek, J.; Sej{\'a}k, M.; and Konop{\'\i}k, M. 2021.
\newblock Czert--Czech BERT-like Model for Language Representation.
\newblock \emph{arXiv preprint arXiv:2103.13031}.

\bibitem[{Song et~al.(2020)Song, Tan, Qin, Lu, and Liu}]{song2020mpnet}
Song, K.; Tan, X.; Qin, T.; Lu, J.; and Liu, T.-Y. 2020.
\newblock MPNet: Masked and Permuted Pre-training for Language Understanding.
\newblock \emph{NeurIPS}.

\bibitem[{Straka et~al.(2021)Straka, N{\'a}plava, Strakov{\'a}, and Samuel}]{straka2021robeczech}
Straka, M.; N{\'a}plava, J.; Strakov{\'a}, J.; and Samuel, D. 2021.
\newblock RobeCzech: Czech RoBERTa, a monolingual contextualized language representation model.
\newblock In \emph{TSD}.

\bibitem[{Svoboda and Brychc{\'\i}n(2018)}]{svoboda2018czech}
Svoboda, L.; and Brychc{\'\i}n, T. 2018.
\newblock Czech dataset for semantic textual similarity.
\newblock In \emph{TSD}. Springer.

\bibitem[{Wang, Reimers, and Gurevych(2021)}]{wang2021tsdae}
Wang, K.; Reimers, N.; and Gurevych, I. 2021.
\newblock {TSDAE}: Using Transformer-based Sequential Denoising Auto-Encoderfor Unsupervised Sentence Embedding Learning.
\newblock In \emph{EMNLP}.

\bibitem[{Wang et~al.(2023)Wang, Yang, Huang, Jiao, Yang, Jiang, Majumder, and Wei}]{wang2022simlm}
Wang, L.; Yang, N.; Huang, X.; Jiao, B.; Yang, L.; Jiang, D.; Majumder, R.; and Wei, F. 2023.
\newblock {S}im{LM}: Pre-training with Representation Bottleneck for Dense Passage Retrieval.
\newblock In \emph{ACL}.

\bibitem[{Wu et~al.(2022)Wu, Gao, Lin, Han, Wang, and Hu}]{wu2022infocse}
Wu, X.; Gao, C.; Lin, Z.; Han, J.; Wang, Z.; and Hu, S. 2022.
\newblock {I}nfo{CSE}: Information-aggregated Contrastive Learning of Sentence Embeddings.
\newblock In \emph{EMNLP}.

\bibitem[{Xiao et~al.(2022)Xiao, Liu, Shao, and Cao}]{liu2022retromae}
Xiao, S.; Liu, Z.; Shao, Y.; and Cao, Z. 2022.
\newblock {R}etro{MAE}: Pre-Training Retrieval-oriented Language Models Via Masked Auto-Encoder.
\newblock In \emph{EMNLP}.

\bibitem[{Zhao et~al.(2022)Zhao, Liu, Ren, and Wen}]{zhao2022dense}
Zhao, W.~X.; Liu, J.; Ren, R.; and Wen, J.-R. 2022.
\newblock Dense Text Retrieval based on Pretrained Language Models: A Survey.
\newblock \emph{arXiv preprint arXiv:2211.14876}.

\bibitem[{Zhuang et~al.(2021)Zhuang, Wayne, Ya, and Jun}]{liu2019roberta}
Zhuang, L.; Wayne, L.; Ya, S.; and Jun, Z. 2021.
\newblock A Robustly Optimized {BERT} Pre-training Approach with Post-training.
\newblock In \emph{CCL}.

\end{thebibliography}

\end{document}